\documentclass[letterpaper, 10 pt, conference]{ieeeconf}  % Comment this line out if 4you need a4paper

\IEEEoverridecommandlockouts                              % This command is only needed if 
\usepackage{hyperref}
\hypersetup{
    colorlinks=true,
    linkcolor=black,    
    urlcolor=blue,
    }
\usepackage{graphicx}
\usepackage[dvipsnames]{xcolor}
\usepackage{booktabs}
\usepackage{epsfig} % for postscript graphics files
\usepackage{amsmath} % assumes amsmath package installed
\usepackage{amssymb}  % assumes amsmath package installed
\usepackage{mathrsfs}
\usepackage{cleveref}
\usepackage{color}
\usepackage{calc}
\usepackage{dsfont}
\usepackage{bm}
\usepackage{float}
\usepackage{multirow}
\newtheorem{theorem}{Theorem}

\newtheorem{problem}{Problem}

\usepackage{xspace}

\usepackage{balance}

\usepackage{etoolbox}
\makeatletter
\patchcmd{\@makecaption}
  {\scshape}
  {}
  {}
  {}
\makeatletter
\patchcmd{\@makecaption}
  {\\}
  {.\ }
  {}
  {}
\makeatother

\newcommand{\param}{\ensuremath{\bm{\theta}}\xspace}
\newcommand{\paramSet}{\ensuremath{\Theta}\xspace}
\newcommand{\paramTrue}{\ensuremath{\bm{\theta^*}}\xspace}
\newcommand{\paramEst}{\ensuremath{\bm{\hat{\theta}}}\xspace}

\newcommand{\paramErrorMax}{\ensuremath{\bm{\tilde{\vartheta}}}\xspace}

\DeclareMathOperator*{\argmin}{arg\,min}

\usepackage{amsfonts} 
\title{\LARGE \bf
Robust Adaptive Safe Robotic Grasping with Tactile Sensing}

\author{Yitaek Kim$^{1}$, Jeeseop Kim$^{2}$, Albert H. Li$^{2}$,  Aaron D. Ames$^{2}$ and Christoffer Sloth$^{1}$ % <-this % stops a space
\thanks{$^{1}$Authors are with the Maersk Mc-Kinney Moller Institute, University of Southern Denmark, Denmark {\tt\small \{yik,chsl\}@mmmi.sdu.dk}}%
\thanks{$^{2}$Authors are with the Departments of \textit{Mechanical and Civil Engineering} and \textit{Computing and Mathematical Sciences}, California Institute of Technology, Pasadena, CA 91125, USA,
 {\tt\small \{jeeseop, alberthli, ames\}@caltech.edu}}
}

\usepackage{tikz}

\newcommand\submittedtext{%
  \footnotesize \textcopyright \text{ }2025 IEEE.  Personal use of this material is permitted.  Permission from IEEE must be obtained for all other uses, in any current or future media, including reprinting/republishing this material for advertising or promotional purposes, creating new collective works, for resale or redistribution to servers or lists, or reuse of any copyrighted component of this work in other works.}
  
\newcommand\submittednotice{%
\begin{tikzpicture}[remember picture,overlay]
\node[anchor=south,yshift=10pt] at (current page.south) {\fbox{\parbox{\dimexpr\textwidth-\fboxsep-\fboxrule\relax}{\submittedtext}}};
\end{tikzpicture}%
}

\begin{document}
\maketitle
\submittednotice
\thispagestyle{empty}
\pagestyle{empty}

%%%%%%%%%%%%%%%%%%%%%%%%%%%%%%%%%%%%%%%%%%%%%%%%%%%%%%%%%%%%%%%%%%%%%%%%%%%%%%%%

\begin{abstract}
Robotic grasping requires safe force interaction to prevent a grasped object from being damaged or slipping out of the hand. In this vein, this paper proposes an integrated framework for grasping with \textit{formal safety guarantees} based on Control Barrier Functions. We first design contact force and force closure constraints, which are enforced by a safety filter to accomplish safe grasping with finger force control. For sensory feedback, we develop a technique to estimate contact point, force, and torque from tactile sensors at each finger. We verify the framework with various safety filters in a numerical simulation under a two-finger grasping scenario. We then experimentally validate the framework by grasping multiple objects, including fragile lab glassware, in a real robotic setup, showing that safe grasping can be successfully achieved in the real world. We evaluate the performance of each safety filter in the context of safety violation and conservatism, and find that disturbance observer-based control barrier functions provide superior performance for safety guarantees with minimum conservatism.
\end{abstract}

%%%%%%%%%%%%%%%%%%%%%%%%%%%%%%%%%%%%%%%%%%%%%%%%%%%%%%%%%%%%%%%%%%%%%%%%%%%%%%%%
\section{Introduction}\label{sec:introduction}
The human hand is capable of adapting to a wide range of complex objects and performing different tasks in daily life robustly \cite{Liu2014}, since it allows smooth and safe force interaction \cite{Tong2024}. Inspired by this, robotic grasping has been investigated across many human-oriented applications such as industrial assembly \cite{LiTME2019}, packing of groceries \cite{ChenRoboSoft2024}, and household tasks \cite{Jessica2024}. Despite the significant progress of robotic grasping over the past decades, the grasping research still revolves around determining proper finger postures for grasping from a kinematic perspective \cite{Zhao2024ICRA}\cite{Albert2023IROS}. At the same time, contact dynamics are often overlooked in existing grasping methods, which are not sufficient for achieving dexterous grasping \cite{TangICRA2024}. This limitation prevents robotic grasping from being applied to more general tasks that require safe behaviors, as they cannot prevent slippage between the hand and object. % \cite{Zeng2023}

To mitigate this limitation, some researchers have focused on force-based grasping techniques. In \cite{ZengIROS2021}, they propose an adaptive force control framework that allows the hand to be compliant by mapping the human hand posture data into the desired force commands. To prevent undesired slippage, \cite{CostanzoICRA2020} proposes a robust force controller for grasping to maintain secure grasping by introducing force feedback.
Similarly, \cite{VanTRO2018} demonstrates a low-level impedance-based controller that incorporated task-based search strategies, comparing its effectiveness on a peg-in-hole task using various robotic hands. Furthermore, a method for regulating the grasping force based on tactile sensors is proposed in \cite{Deng2020} to increase grasping stability for unknown objects by leveraging data-driven models like deep neural networks or Gaussian mixture models to detect and estimate contact status and force. 
Others have demonstrated that online adaptive control strategies can account for disturbances and model error \cite{Khadivar2023TRO} using dynamical systems-based \cite{Billard2011DS} approaches for dexterous manipulation.

The modeling of contact dynamics has also played a key role in many robust grasping works. In \cite{Pfanne2020RAL}, internal and friction forces of a grasped object are determined in order to provide stable finger gaits so that the proposed impedance controller can handle the dynamic changes in object state during reconfiguration of the fingers. Similarly, \cite{Xue2023IROS} proposes a hybrid position/force-controller used to generate a finger gaiting sequence while considering dynamic constraints imposed by friction cones, enabling stable control over object pose. 

\begin{figure}[t]
    \centering
    \includegraphics[width=1\linewidth]{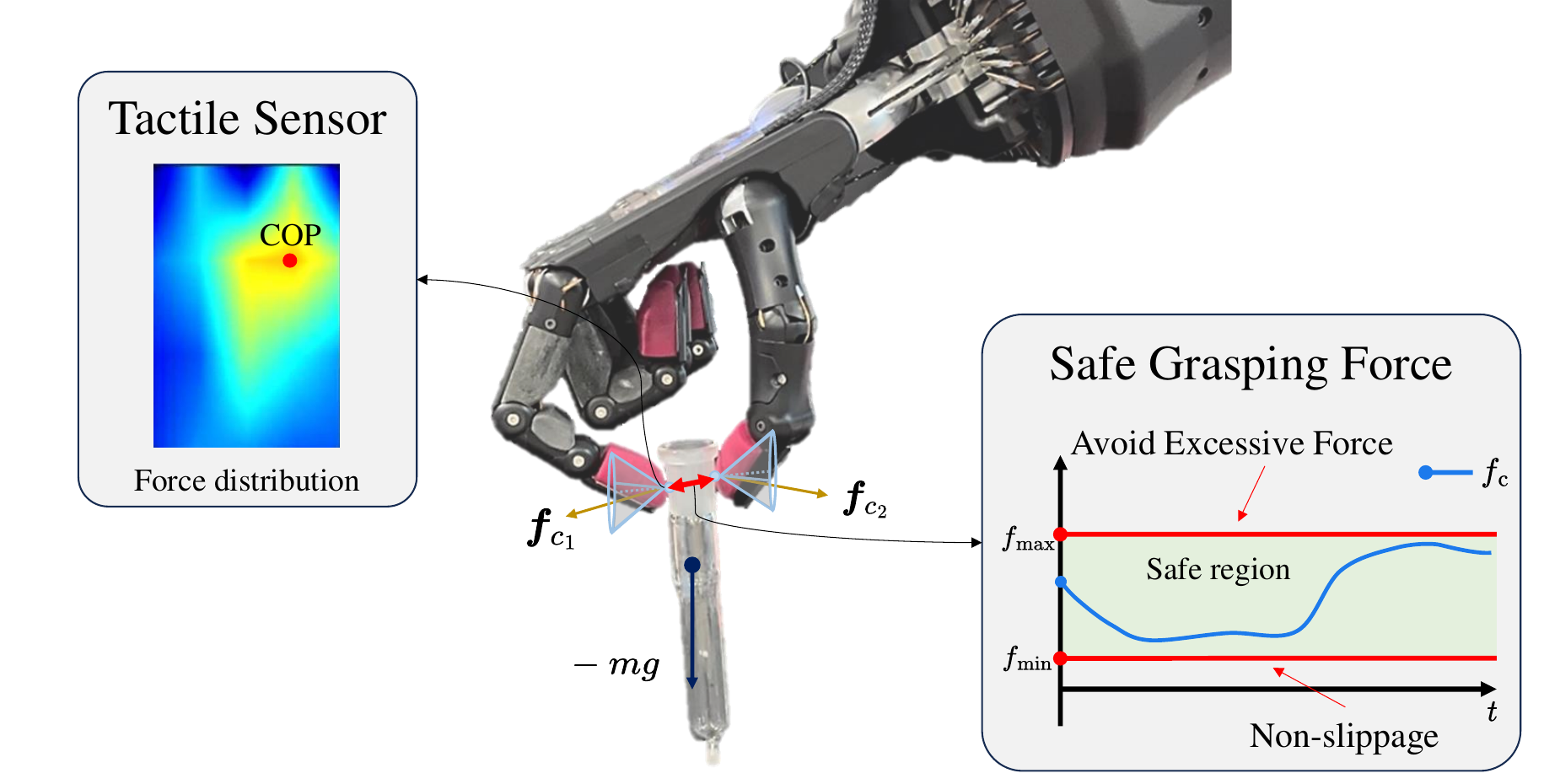}
    \caption{Safe robotic grasping for fragile lab glassware.}
    \vspace{-0.6cm}
    \label{fig:fingers}
\end{figure}

Such force-based grasping methods have primarily focused only on stable grasping, neglecting safety considerations, which are crucial in real-world applications. For instance, safety regulations are essential for laboratory automation \cite{WOLF202218} and space applications including robotic hands \cite{Hadi2024}. Therefore, ensuring safety in robotic grasping remains an open challenge beyond merely achieving compliant manipulation. To that end, a few works exist, proposing methods such as safe grasping with impact force measurements \cite{Mavrakis2017IROS}, safety-optimized strategies with visual depth data and false-positive detection for human safety \cite{Li2024UR}, and an integrated framework for predicting safe grasping forces using transformers \cite{Han2024ASME}.

\begin{figure*}[t]
    \centering
    \includegraphics[width=1\linewidth]{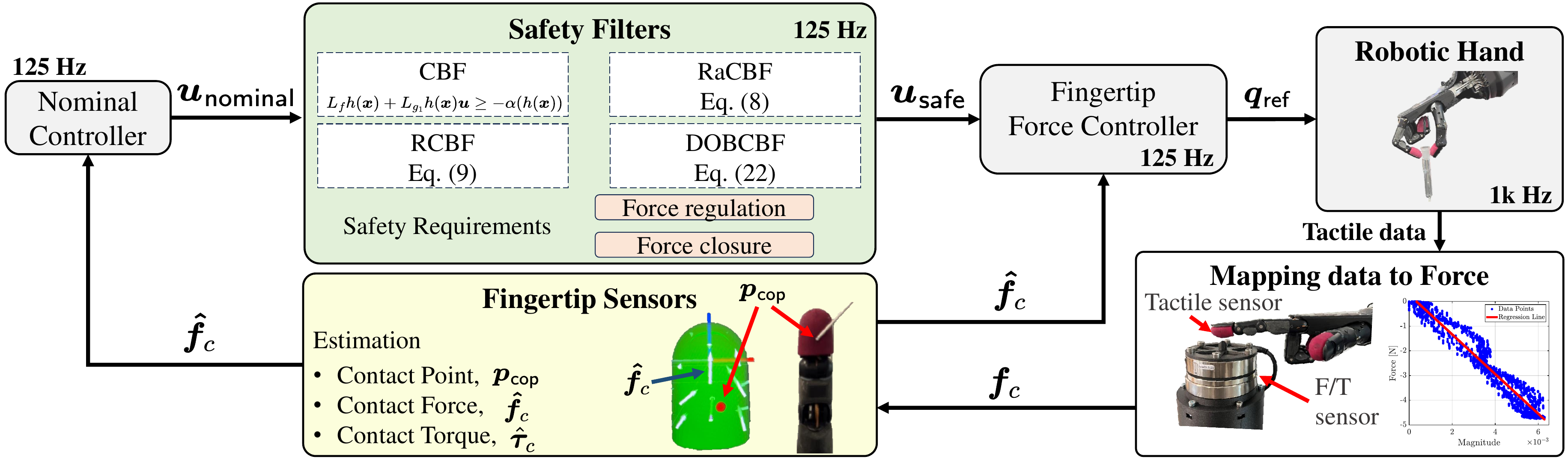}
    \caption{Overview of the proposed safe grasping framework. The framework consists of three main components: safety filters, estimation of contact information, and fingertip force controller. Safety filters include CBF, RaCBF, RCBF, and DOBCBF with a disturbance observer. Tactile sensor data from the hand is mapped to the actual force, and the contact force/torque and point on a fingertip are estimated to be used in the controllers. The fingertip force controller is designed to track safe control input to the hand.}
    \vspace{-0.4cm}
    \label{pro:framework}
\end{figure*}

Nevertheless, to the best of our knowledge, few studies take into account \textit{formal safety guarantees} (e.g., force regulation and closure to avoid slippage) for force-based grasping control; thus, imbuing grasping algorithms with formal safety guarantees should be investigated for broader use cases such as medical, space, and chemical laboratory applications. Safety guarantees can be efficiently achieved by using reachability analysis \cite{Fisac2019}, set invariance techniques \cite{Wolff2004}, and control barrier functions (CBFs) \cite{Ames2019CBFtheoryandapplications}\cite{lopez2020robust}\cite{JANKOVIC2018359}.
In this paper, we focus on CBF-based controllers to ensure safety guarantees in robotic grasping. CBF-based controllers enforce the forward invariance of a defined safe set and can be efficiently implemented using quadratic programming, enabling their widespread use in many applications \cite{dawson2022barrier}\cite{YKECC2024}. 
The use of  CBFs for safe grasping was first proposed by \cite{Shaw2021}. Their methodology used robust CBFs in order to handle external disturbances, and consequently, achieved safe grasping while avoiding undesired slippage. 

\subsection{Contributions}
The main novelty presented in this paper is the introduction of an integrated robotic safe grasping framework with tactile sensing as shown in Fig.~\ref{pro:framework}. The framework is specifically designed to maintain maximum and minimum safe grasping forces and achieve safe force closure by employing robust and adaptive approaches in the presence of model uncertainty.
The contributions of this paper are outlined as follows: 
\begin{itemize}
    \item We present a safe grasping framework with \textit{formal safety guarantees}, based on CBFs schemes as safety filters. The scalability of the framework presented in this paper is demonstrated by applying several safety filters, including CBF \cite{Ames2019CBFtheoryandapplications}, robust adaptive CBF \cite{lopez2020robust} (RaCBF), robust CBF (RCBF) \cite{JANKOVIC2018359}, and disturbance observer CBF (DOBCBF) \cite{Wang2023}.  
    \item To this end, we implement force control with an inner position loop for a dexterous robot hand and develop sensor processing techniques to estimate contact force and point from electromagnetic tactile sensors on fingertips. This approach enables us to use contact dynamics model effectively in CBFs-based controller.
    \item The effectiveness of our framework is empirically validated with multiple objects in a scenario where the grasping force must be regulated within safety limits to prevent damage to objects and slippage as well. Additionally, we provide the detailed comparisons between each safety filter in the context of safety violation and conservatism.  We use Shadow Robot platform for validations as shown in Fig.~\ref{fig:fingers}.
    \item To the best of our knowledge, this work is the first attempt not only to introduce an integrated framework with tactile sensing for grasping with \textit{formal safety guarantees}, but also to demonstrate it in the real-world. 
\end{itemize}

We show significant improvements in our framework compared to \cite{Shaw2021} with respect to adaptation, conservatism, and grasping force regulation. The proposed framework incorporates adaptation to parametric uncertainty while estimating external disturbances, making it less conservative than \cite{Shaw2021}. Additionally, since \cite{Shaw2021} maintains a strong grasping force to prevent slipping without regulating the force, it may lead to applying excessive force and potentially damaging the grasped object. In this work, constraints including contact force and force closure are simultaneously considered.

This paper is organized as follows. Section~\ref{sec:problem_formulation} defines a problem solved in the paper and Section~\ref{sec:preliminary_knowledges} provides safety filters and a disturbance observer as preliminaries. Section~\ref{sec:method} presents the proposed framework. Subsequently, we verify our framework in numerical simulations in Section~\ref{sec:simulation} and validate it in the real multiple experiments in Section~\ref{sec:experimental_validation}. Lastly, we conclude our paper in Section~\ref{sec:conclusions}.

\section{Problem Formulation}\label{sec:problem_formulation}
The framework for safe grasping is based on Control Barrier Functions (CBFs). The success of a grasp depends on the contact dynamics between fingers and objects; however, this is often unknown. Therefore, we consider the following control affine systems to represent contact force dynamics with parametric uncertainty or uncertainty given by an unknown additive disturbance: 
\begin{align}
        \dot{\bm{x}} &= f(\bm{x}) + F(\bm{x})\paramTrue+g_1(\bm{x})\bm{u},
        \label{prob:sys_linearized_model}\\
        \dot{\bm{x}} &= f(\bm{x}) + g_1(\bm{x})\bm{u} + g_2(\bm{x})\bm{d}, 
        \label{prob:sys_disturbance_model}
\end{align}
where $\bm{x} \in \mathcal{X} \subset \mathbb{R}^n$ is the system state, $f: \mathcal{X} \rightarrow \mathbb{R}^n$, $g_1: \mathcal{X} \rightarrow \mathbb{R}^{n\times m}$, and $g_2: \mathcal{X} \rightarrow \mathbb{R}^{n\times q}$ are locally Lipschitz continuous functions, $F:\mathcal{X}\rightarrow \mathbb{R}^{n \times k}$ is a smooth function with $F(\bm{0})=\bm{0}$, $\paramTrue \in \paramSet \subset \mathbb{R}^k$ is the vector of constant unknown parameters, and $\bm{u} \in \mathcal{U} \subset \mathbb{R}^m$ represents control force input, and $\bm{d} \in \mathbb{R}^q, ||\bm{d}|| \leq w_0, ||\dot{\bm{d}}|| \leq w_1$ is a bounded external disturbance. 
We address the following problem:
\begin{problem}
Design a framework that enables safe grasping, i.e., by ensuring satisfaction of contact force and force closure constraints.
\end{problem}

\section{Preliminaries}\label{sec:preliminary_knowledges}
This section briefly introduces the contact dynamics corresponding to \eqref{prob:sys_linearized_model} and \eqref{prob:sys_disturbance_model}, the core concept of safety filters, and a nonlinear disturbance observer.

\subsection{Contact Dynamics Models}
Consider the following second-order Kelvin-Voigt model \cite{kelvinModel} for contact forces:
\begin{equation}
     \bm{f}_{c}(\bm{x}) = -diag(\bm{k})\bm{p}-diag(\bm{b})\dot{\bm{p}} \label{kv_contact_model}
\end{equation}
where $\bm{x}= \begin{bmatrix} {\bm{p}} &\dot{\bm{p}}\end{bmatrix}^{\top}$ is the system state, $\bm{p} \in \mathbb{R}^n_+$ is the penetration of the contact surface, $\bm{f}_c = \{f_{\textnormal{cx}},f_{\textnormal{cy}},f_{\textnormal{cz}}\}^{\top}$ is the contact force, $diag(\cdot)$ is a square matrix with the vector elements on the diagonal, and $\bm{k}\in \mathbb{R}^n_{>0}$, and $\bm{b}\in \mathbb{R}^n_{>0} $ are stiffness and damping parameters, respectively. From \eqref{kv_contact_model}, we represent the following state-space model corresponding to \eqref{prob:sys_linearized_model}:
\begin{equation}
\dot{\bm{x}} = \underbrace{\begin{bmatrix}
\dot{\bm{p}} \\
\bm{0}
\end{bmatrix}}_{f(\bm{x})} -\frac{1}{m_o}\underbrace{\begin{bmatrix}\bm{0}&\bm{0}\\ diag(\bm{p}) & diag(\dot{\bm{p}})\end{bmatrix}}_{F(\bm{x})}\underbrace{\begin{bmatrix}
\bm{k} \\
\bm{b}
\end{bmatrix}}_{\theta^*}+\underbrace{\begin{bmatrix}
\bm{0}\\
\frac{1}{m_o}\bm{I}_m
\end{bmatrix}}_{g_1(\bm{x})}\bm{u}, \label{contact_system_model_param}
\end{equation}
where $m_o\in \mathbb{R}_{>0}$ is mass, and $\bm{u}\in \mathbb{R}^m$ is the control force from each finger. $\bm{I}_m\in\mathbb{R}^{m\times m}$ represents an $m\times m$ dimensional identity matrix where $n=m$ in this case.
For the system \eqref{prob:sys_disturbance_model}, we employ the following model:
\begin{align}
    f(\bm{x}) &= \begin{bmatrix}
\dot{\bm{p}} \\
\frac{-1}{m_o}\big(diag(\bm{p})\bm{k} + diag(\dot{\bm{p}})\bm{b}\big)
\end{bmatrix}, \nonumber \\
g_1(\bm{x}) &= \begin{bmatrix}
\bm{0}\\
\frac{1}{m_o}\bm{I}_m
\end{bmatrix}, \quad g_2(\bm{x}) = \begin{bmatrix}
    \bm{0}\\
    \frac{1}{m_o}\bm{I}_q
\end{bmatrix}, \label{contact_system_model_disturb} 
\end{align}
where $\bm{I}_q\in\mathbb{R}^{n\times m}$ is a disturbance characteristic matrix.

Lastly, the soft finger friction cone model is defined as follows \cite{Prattichizzo2016}:
    \begin{equation}
       \frac{1}{\mu}\sqrt{f^2_{\textnormal{cx}}+f^2_{\textnormal{cy}}} +\frac{1}{a\eta}|\tau_{\textnormal{cz}}|\leq f_{\textnormal{cz}}, \label{pre:friction_model}
    \end{equation}
where  $f_{\textnormal{cx}}$ and $f_{\textnormal{cy}}$ are tangential forces on the contact surface; $\mu \in \mathbb{R}_+$ is the friction coefficient; ${\tau}_{\textnormal{cz}} \in \mathbb{R}$ is the torsional friction; $\eta$ is the torsional friction coefficient; and $a$ is an auxiliary parameter that ensures consistent units.

\subsection{Robust Adaptive Control Barrier Functions}
Consider the system \eqref{prob:sys_linearized_model} and a parameterized safe set with the parameter $\param$, $\mathcal{S}_{\param} = \{ \bm{x}\in\mathcal{X}|h_r(\bm{x},\param) \geq 0 , \, \param \in \paramSet \}$,
where  $h_r: \mathcal{X} \times \paramSet \rightarrow \mathbb{R}$ is a continuously differentiable function. Due to parametric uncertainty, we define a robust and tightened safe set $\mathcal{S}^r_{\param} \subseteq \mathcal{S}_{\param}$ as:
    \begin{align}
        \mathcal{S}_{\param}^{r} = \left\{ \bm{x}\in\mathcal{X}:h_r(\bm{x},\param)\geq\frac{1}{2}\paramErrorMax^{\top}\Gamma^{-1}\paramErrorMax, \param \in \paramSet  \right\}, \label{pre:racbf_safe_set}
    \end{align}
        where $\paramErrorMax \in \paramSet$ is the maximum possible estimation error between $\paramTrue$ and $\paramEst$ with respect to the gain defined by $\Gamma^{-1}$.
    \begin{theorem}[\cite{lopez2020robust}]
        The function $h_r$ is a Robust adaptive Control Barrier Function (RaCBF) for \eqref{prob:sys_linearized_model} if there exists an extended class $\mathcal{K}_\infty$ function $\alpha(\cdot)$ such that
        \begin{align}\label{racbf_constraint}
            &\sup_{\bm{u}\in\mathcal{U}}
            \Bigg\{
                \frac{\partial h_r}{\partial \bm{x}}(\bm{x}, \paramEst)
                \bigg(
                    f(\bm{x}) + F(\bm{x})\lambda(\bm{x}, \paramEst) + g_1(\bm{x})\bm{u}
                \bigg)
            \Bigg\} \nonumber \\
            &\qquad \qquad \qquad \geq -\alpha\bigg(h_r(\bm{x}, \paramEst)-\frac{1}{2}\paramErrorMax^{\top}\Gamma^{-1}\paramErrorMax\bigg),
        \end{align}
        with $\lambda(x,\param) \triangleq {\param} - \Gamma\Big(\frac{\partial h_r}{\partial \param}(x,\param)\Big)^{\top}$,
        for any $\paramEst \in \paramSet$ satisfying the adaptation law $\dot{\paramEst} =-\Gamma\Big(\frac{\partial h_r}{\partial \bm{x}}(\bm{x},\paramEst)F(\bm{x})\Big)^{\top}$, 
        and where $\Gamma \in \mathbb{R}^{n \times n}$ is an adaptive gain whose minimum eigenvalue satisfies $\lambda_{\text{min}}(\Gamma) \geq \frac{\vert\vert \paramErrorMax \vert\vert^2}{2h_r(x,\param)}$. The system \eqref{prob:sys_linearized_model} is safe with respect to \eqref{pre:racbf_safe_set} if $h_r$ is a RaCBF. 
        \label{pre:racbf_theorem}
    \end{theorem}

\subsection{Robust Control Barrier Functions}
Consider the system \eqref{prob:sys_disturbance_model} and the safe set $\mathcal{S} = \{\bm{x} \in \mathcal{X} \text{ }\vert\text{ } h(\bm{x}) \geq 0\}\label{pre:safe_set}$ defined as the 0-superlevel set of a continuously differentiable function $h: \mathbb{R}^n \rightarrow \mathbb{R}$.
    \begin{theorem}[\cite{JANKOVIC2018359}]
        The function $h$ is a Robust Control Barrier Function (RCBF) if there exists a control input $\bm{u}$ and an extended class $\mathcal{K}_{\infty}$ function $\alpha$ such that for all $\bm{x}\in \mathcal{X}$,
        \begin{equation}
            \sup_{\bm{u}\in\mathcal{U}}[L_{f}h(\bm{x}) + L_{g_1}h(\bm{x})\bm{u} -||L_{g_2}h(\bm{x})||w_0 + \alpha\big(h(\bm{x})\big)] \geq 0.
            \label{rcbf_def}
        \end{equation}
    The system \eqref{prob:sys_disturbance_model} is safe with respect to $\mathcal{S}$ if $h$ is a RCBF.
        \label{rcbf_theorem}
    \end{theorem}

\subsection{Nonlinear Disturbance Observer (NDOB)}
Consider the uncertain system \eqref{prob:sys_disturbance_model}. Assuming that $||\bm{d}(t)|| \leq w_0$ and $||\dot{\bm{d}}(t)|| \leq w_1$, we deploy the following nonlinear disturbance observer proposed by \cite{DOBChens2004}:
\begin{align}
    \hat{\bm{d}}(t) &= \bm{z} + \alpha_d P(\bm{x})  \label{pre:est_dob} \\
    \dot{\bm{z}}(t) &= -\alpha_d L_d\Big(f(\bm{x}) + g_1(\bm{x})\bm{u} + g_2(\bm{x})\hat{\bm{d}}(t)\Big), \nonumber 
\end{align}
where $\alpha_d$ is a constant positive adjustable parameter, and $P(\bm{x})$ is the gain function for the estimator satisfying $\frac{\partial P}{\partial \bm{x}} = L_d(\bm{x})$. $L_d(\bm{x})$ is the predefined gain function satisfying $\bm{e}_d^{\top}\bm{e}_d \leq \bm{e}_d^{\top}L_d(\bm{x})g_2(\bm{x})\bm{e}_d$ \cite{Mohammadi}, where $\bm{e}_d = \bm{d} -\hat{\bm{d}}$ is the estimation error, and $\bm{e}_d$ is uniformly bounded by $||\bm{e}_d(t)|| \leq \sqrt{\frac{2ck||\bm{e}_d(0)||^2e^{-2kt}+w^2_1(1-e^{-2kt})}{2ck}}:= M_d$, where $k = \alpha_d - \frac{c}{2}$, and $c$ is a non-negative parameter that satisfies $c<2\alpha_d$.

\section{The Proposed Framework}\label{sec:method}
In this section, we present a comprehensive framework for safe robotic grasping based on CBF-based controllers. The main components of our framework consist of safety filters, a fingertip contact force controller, and the estimation of contact points and forces/torques,  as illustrated in Fig.~\ref{pro:framework}. For the contact dynamics \eqref{prob:sys_linearized_model} and \eqref{prob:sys_disturbance_model}, we formulate safety constraints for grasping forces which achieve safe grasping. To track the reference control input from safety filters, we use a joint position-based force controller for each fingertip. Additionally, we estimate contact points and force/torque using tactile sensors embedded on the fingertips. The proposed framework is developed on Robot Operating System (ROS) for extensive use.

\subsection{Safety-Critical Grasping}
Based on \textit{Problem~1}, we define two contact force constraints: regulating contact force to prevent excessive force and ensuring force closure to avoid slippage. 
\subsubsection{Contact Force constraint}
To maintain the contact force $\bm{f}_c = \{f_{\textnormal{cx}},f_{\textnormal{cy}},f_{\textnormal{cz}}\}^{\top}$ within admissible safety limits during grasping, we define the safe force range as follows:
\begin{equation}
    \bm{f}_{\textnormal{min}} \leq \bm{f}_{c} \leq \bm{f}_{\textnormal{max}},
    \label{pro:const1}
\end{equation}
where $\bm{f}_{c}(\bm{x}) = -diag(\bm{k})\bm{p}-diag(\bm{b})\dot{\bm{p}}$ is a contact force, and  $\bm{f}_{\textnormal{max}} \in \mathbb{R}^3$ and $\bm{f}_{\textnormal{min}} \in \mathbb{R}^3$ are the vectors of the admissible maximum and minimum contact forces, respectively. We define the sets of the constraint \eqref{pro:const1} as: 
\begin{equation}
        \mathcal{H}_{f_{\textnormal{c}}} = \mathcal{H}_{f_{\textnormal{min}}}  \cap \mathcal{H}_{f_{\textnormal{max}}}, \label{pro:contact_force_const1}
\end{equation}
where
\begin{align}
    \mathcal{H}_{f_{\textnormal{min}}} &=\{ \bm{x}\text{ }|\text{ } \bm{f}_{c}(\bm{x}) - \bm{f}_{\textnormal{min}} \geq 0\}, \label{force_min_lim}\\ 
    \mathcal{H}_{f_{\textnormal{max}}} &=\{ \bm{x}\text{ }| \text{ } -\bm{f}_{c}(\bm{x}) + \bm{f}_{\textnormal{max}} \geq 0\}. \label{force_max_lim}
\end{align}

\subsubsection{Force Closure}
To prevent slipping during grasping, we first consider a Coulomb friction model given in \eqref{pre:friction_model} and linearize it to be conservatively approximated in a set of tangential planes to the original cone as follows \cite{analysishandsJeffrey1986}:
\begin{align}
    \Lambda(\mu,a,\eta)\mathcal{F}^C_c &\geq 0, \\
    \Lambda(\mu,a,\eta) &= \begin{bmatrix}
        1 & 0 & \mu & \frac{1}{a\eta}\\
        -1 & 0 & \mu & -\frac{1}{a\eta} \\
        0 & 1 & \mu & \frac{1}{a\eta}\\
        0 & -1 & \mu & -\frac{1}{a\eta}\\
    \end{bmatrix},
\end{align}
where $\mathcal{F}^C_c = \{f_{\textnormal{cx}},f_{\textnormal{cy}},f_{\textnormal{cz}},|\tau_{\textnormal{cz}}|\}^{\top}$ is the vector of contact forces and normal moment relative to the contact frame. We define the safe set for the force closure constraint as:
\begin{equation}
    \mathcal{H}_s =\{ \mathcal{F}^C_c\in \mathbb{R}^4| \Lambda(\mu,a,\eta)\mathcal{F}^C_c \geq 0\}. \label{force_closure_const}
\end{equation}
Consequently, we obtain the following intersection of constraint sets for safe grasping:
\begin{equation}
    \mathcal{H} = \mathcal{H}_{f_{\textnormal{c}}}  \cap \mathcal{H}_s, \label{all_constraints}
\end{equation}
assuming $\mathcal{H}$ is non-empty.

\subsection{Safety Filters for Safe Grasping}
When given $\mathcal{H}$, the safe control input is provided to minimize the following objective function:
\begin{equation}
     \bm{u}_{\mathsf{safe}} = \quad \argmin_{\bm{u}\in\mathcal{U}} \frac{1}{2} {\vert\vert \bm{u} - \bm{u}_{\mathsf{nominal}} \vert\vert}^2,
\end{equation}
where $\bm{u}_{\mathsf{nominal}} \in \mathbb{R}^m$ is a nominal control input, and $\bm{u}._{\mathsf{safe}}\in \mathbb{R}^m$ is a safe control input from each safety filter. The constraint \eqref{all_constraints} is easily enforced by a standard CBF \cite{Ames2019CBFtheoryandapplications} for the system \eqref{prob:sys_disturbance_model} without considering $\bm{d}$. However, since standard CBFs are vulnerable to model uncertainties, we use following methods in our framework: RaCBFs \cite{lopez2020robust}, RCBFs \cite{JANKOVIC2018359}, and DOBCBFs \cite{Wang2023}. RaCBFs resolve inherent parametric model uncertainty for the system \eqref{contact_system_model_param}, and can provide a safe input by imposing $\mathcal{H}$ into \eqref{racbf_constraint} with $\paramErrorMax$. For more general use cases, RCBFs for the system \eqref{contact_system_model_disturb} can attenuate bounded external disturbances by applying $\mathcal{H}$ into \eqref{rcbf_def} with $w_0$. Lastly, the disturbance observer \eqref{pre:est_dob} can estimate the external disturbance in the system \eqref{prob:sys_disturbance_model}. $P(\bm{x})$ and $L_d$ are problem-specific gain functions that we choose to be:
\begin{align}
    P(\bm{x}) &= diag(\bm{k})\bm{p} + diag(\bm{b})\dot{\bm{p}}, \\
    L_d &= \begin{bmatrix}
        \bm{k}^{\top} & \bm{b}^{\top}
    \end{bmatrix}.
\end{align}
Subsequently, we include the estimated disturbance into safety condition for DOBCBFs \cite{DOBChens2004} as follows:
    \begin{equation}
     \psi_0 + L_{g_1}h(\bm{x})\bm{u} \geq 0,
            \label{dob_def}
    \end{equation}
    with $\psi_0 = L_{f}h(\bm{x}) + L_{g_2}h(\bm{x})\hat{\bm{d}}-\frac{w^2_1}{2c\beta}-\frac{\beta||L_{g_2}h||^2}{4\nu-2c-2\alpha} +\alpha(h(\bm{x}))$ where the positive constant $\beta, \nu$ such that $\nu>\frac{\alpha+c}{2}$ and $\beta > \frac{||\bm{e}_d(0)||^2}{2h(\bm{x}(0))}$.
In the subsequent section, a fingertip force controller for the robotic hand is designed to track a safe input generated by each safety filter.

\subsection{Fingertip Contact Force Controller}
We use a joint position-based force controller to track the desired force reference.
%, since our real robot setup, Shadow Robot only supports joint position commands.
For the control force input, $\bm{u}$ is defined as: 
\begin{equation}
    \bm{u} = K_p\tilde{\bm{f}}, \label{nominal_control_force}
\end{equation}
where $K_p \in \mathbb{R}$ is a non-negative proportional gain, and $\tilde{\bm{f}} \in \mathbb{R}^3$ is the error between the desired contact force $\bm{f}_{\textnormal{d}}$ and the current force $\bm{f}_{\textnormal{c}}$. To generate the control force \eqref{nominal_control_force} in each finger, we use the following force controller for each finger: 
\begin{equation}
    \bm{q}_{\textnormal{ref}} = \bm{q}_{\textnormal{d}} + k_{\textrm{p}}\tilde{\bm{\tau}} + k_{\textrm{i}}\int\tilde{\bm{\tau}}dt   + k_{\textrm{d}}\dot{\tilde{\bm{\tau}}} \label{pro:jointController},
\end{equation}
where $\bm{q} \in \mathbb{R}^r$ is the current joint position and $\bm{q}_{\textnormal{ref}} \in \mathbb{R}^r$, $\bm{q}_{\textnormal{d}} \in \mathbb{R}^r$, $\tilde{\bm{\tau}} = J^{\top}(\bm{q})\tilde{\bm{f}} \in \mathbb{R}^r$ are the joint's reference trajectory, feedforward command, and error torque, respectively. $r$ denotes the DoFs of each finger and $J(\bm{q})$ is the Jacobian, and $k_{\textrm{p}},k_{\textrm{i}},k_{\textrm{d}} \in \mathbb{R}$ are constant controller gains. 

\subsection{Estimating Contact Points, Forces, and Torques}
\begin{figure}[h]
    \centering
\includegraphics[width=1\linewidth]{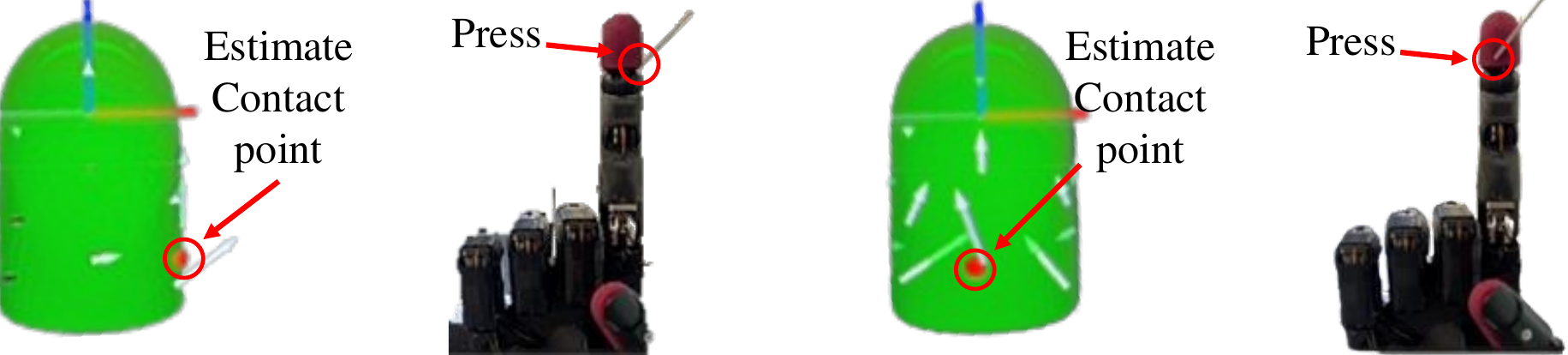}
    \vspace{-0.7cm}
    \caption{The performance of contact point estimation based on tactile sensors}
    \vspace{-0.35cm}
    \label{fig:contact_points}
\end{figure}

For our experiments, we use the Shadow hand, an anthropomorphic hand equipped with seventeen electromagnetic tactile sensors per finger, which provide three-axis magnetic values corresponding to fingertip deformations, allowing us to measure local force readings and contact points. To this end, we design a straightforward experiment to obtain a linear regression model that maps deformations to forces. As shown in Fig.~\ref{pro:framework}, we collect real-world sensor and force data while the hand linearly presses and releases an ATI 6-axis force/torque sensor. Consequently, we obtain the model shown in the bottom right of Fig.~\ref{pro:framework}, which we use in our experiments.
Moreover, because the estimate of contact location is critical for robotic grasping, we develop a novel estimator that infers this location from the center of pressure of the seventeen tactile sensors per finger. Denote by $\bm{s}_i \in \mathbb{R}^3$ and $i=\{1,2,\cdots,17\}$ the sensor values. First, we select the sensor that has the highest magnitude, denoted $\bm{s}^*$ in $\bm{s}_i$. Subsequently, we define a set that includes $s^*$ and its surrounding $k$ sensors: $\mathcal{S}_o = \{\bm{s}^*, \bm{s}_1, \cdots, \bm{s}_k\}.$ Then, the contact point with respect to fingertip frame is calculated via a weighted combination of the sensor values and positions:
\begin{equation}
    \bm{p}_{\textnormal{cop}} = \gamma_*\bm{p}^* + \gamma_1\bm{p}_1 + \cdots + \gamma_k\bm{p}_k, 
\end{equation}
where $\bm{p}_{\textnormal{cop}} \in \mathbb{R}^3$ is the center of pressure relative to the fingertip frame, $\bm{p}^*, \bm{p}_1, \ldots, \bm{p}_k \in \mathbb{R}^3$ are the positions of each sensor, and $\gamma_{*}, \gamma_1, \ldots, \gamma_k$ are the normalized sensor values such that $\gamma_{*} + \gamma_1 + \cdots + \gamma_k = 1$.
Lastly, the contact force $\bm{f}_c$ is approximated by the average of $\mathcal{S}_o$, and the contact torque $\bm{\tau}_c$ is approximated as $\bm{\tau}_c \approx \bm{p}_{\textnormal{cop}}\times \bm{f}_c$. The performance of contact point and force are shown in Fig.~\ref{fig:contact_points}.

%\subsection{Safe Grasping Framework}
% The main components of our framework consist of safety filters, the estimation of contact information, and a fingertip contact force controller as illustrated in Fig.~\ref{pro:framework}. To achieve safe grasping, we select a safety filter depending on a grasping scenario, and estimate the contact point and force/torque using the tactile sensors. To track the desired reference contact force, we implement two controllers: nominal and a fingertip force controller. A proportional controller is used as the nominal controller based on the measured contact force from the sensors. The output of the nominal controller is subsequently filtered by the safety filter and afterwards, the safe control input from the safety filter serves as the reference force in the fingertip controller. To track this safe control input, we utilize a joint position-based force controller \eqref{pro:jointController} for the finger and send $\bm{q}_{\textnormal{ref}}$ to the robot. The proposed framework is developed on Robot Operating System (ROS) for extensive use.

\section{Simulation}\label{sec:simulation}
\begin{figure}[t]
    \centering
\includegraphics[width=1\linewidth]{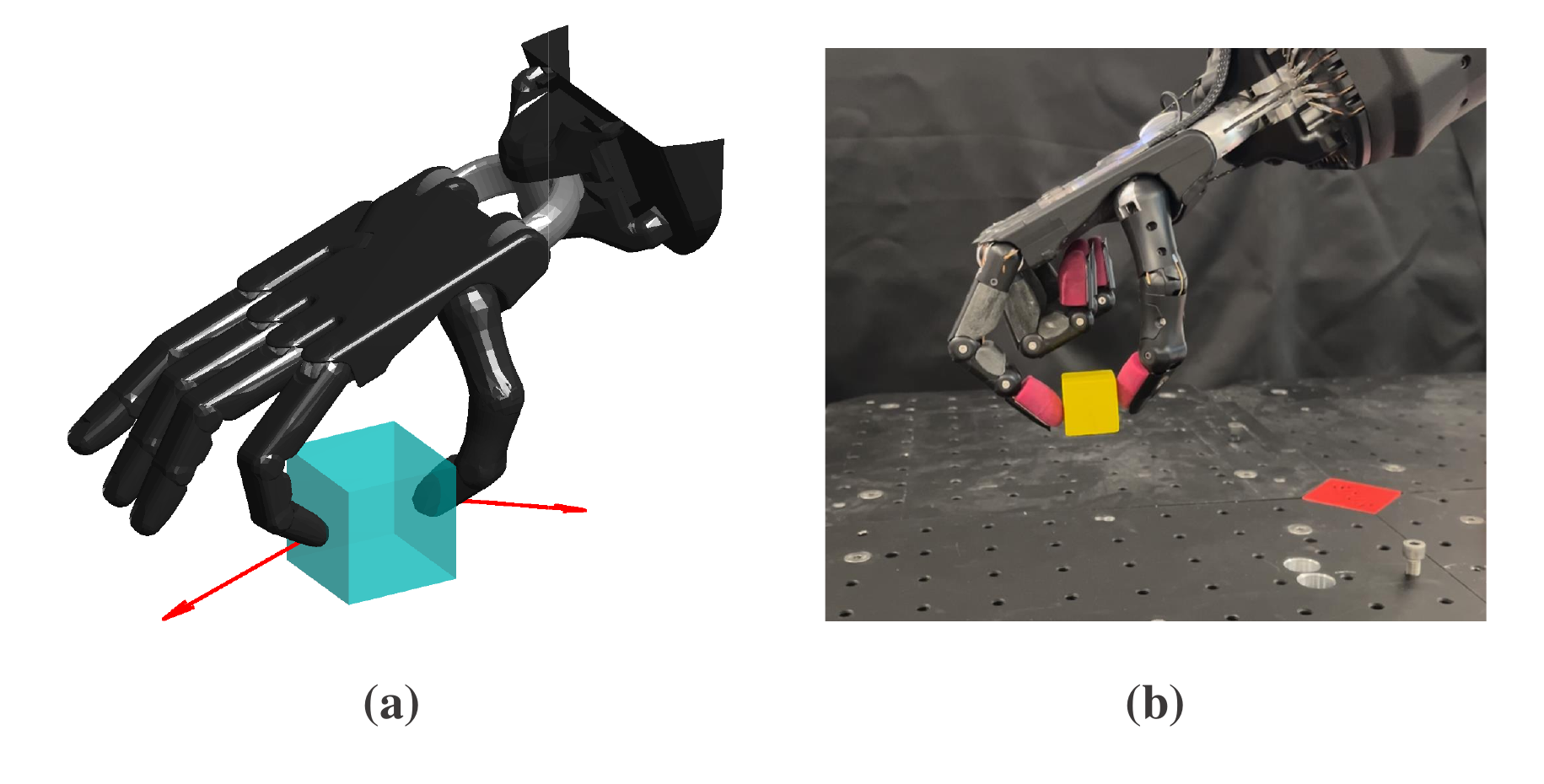}
    \vspace{-0.8cm}
    \caption{(a) and (b) show the simulation and real experiment setups, respectively.}
    \vspace{-0.35cm}
    \label{fig:setups}
\end{figure}

In this section, we conduct numerical simulations using MATLAB, aiming to evaluate the proposed framework for safe grasping in terms of safety violation and conservatism.
% The simulation is carried out under a grasping scenario with two fingers on a shadow Robot hand model
We simulate a two-finger grasping scenario on the Shadow hand as shown in Fig.~\ref{fig:setups} (a). We first generate a linear pinching motion to grasp an object between the front finger and the thumb, and then the feedforward joint trajectory, $\bm{q}_{\textnormal{d}}$ is obtained through inverse kinematics for the front finger. Subsequently, we use the finger force controller \eqref{pro:jointController} to control the front finger to track the desired force.
We assume that the initial grasping condition is within a safe set satisfying the following:
\begin{itemize}
    \item normal reaction forces are equal to or greater than $-6$N.
    \item contact forces lie in a (linearly-approximated) friction cone with a friction coefficient of $\mu$.
\end{itemize}
In this scenario, we use Kelvin-Voigt model as a real contact force model and introduce parametric model uncertainties in both stiffness and damping.
\begin{figure}[ht]
    \centering
\includegraphics[width=1\linewidth]{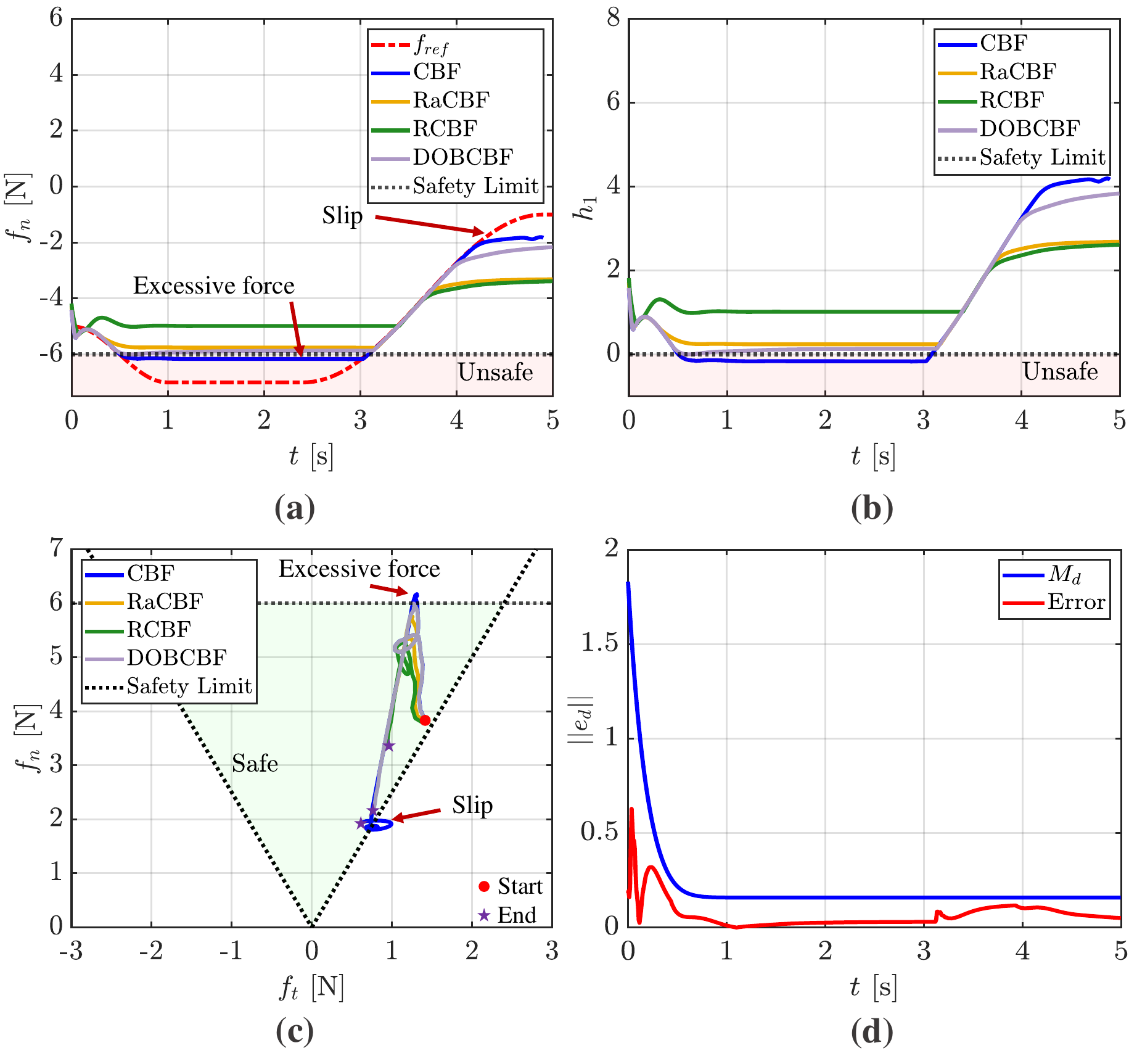}
    \vspace{-0.7cm}
    \caption{(a) shows contact normal force on the contact surface, and (b) presents the control barrier function, $h_1$ which regulates the contact normal force from \eqref{force_min_lim}. (c) shows the contact force is in the friction cone from \eqref{force_closure_const} to achieve force closure. Lastly, (d) plots the disturbance estimation error bound and the actual error.}
    \vspace{0.1cm}
    \label{fig:sim_uncertainty}
\end{figure}

We analyze the performance of each safety filter in terms of safety violations and conservatism. As a baseline, we implement a standard non-robust CBF. As shown in Fig.~\ref{fig:sim_uncertainty} (a) and (b), the standard CBF (blue) violates minimum force regulation and applies excessive force to the object. This occurs because there is a discrepancy between the CBF model and the real model in model parameters. In contrast, the RCBF (green) and RaCBF (orange) show robustness to the induced model uncertainty. The RCBF incorporates robustness into its safety constraints by accounting for model uncertainty as external disturbances. On the other hand, the RaCBF considers the maximum possible parameter error bound in its constraints.

\begin{figure*}[t]
    \centering
    \includegraphics[width=1\linewidth]{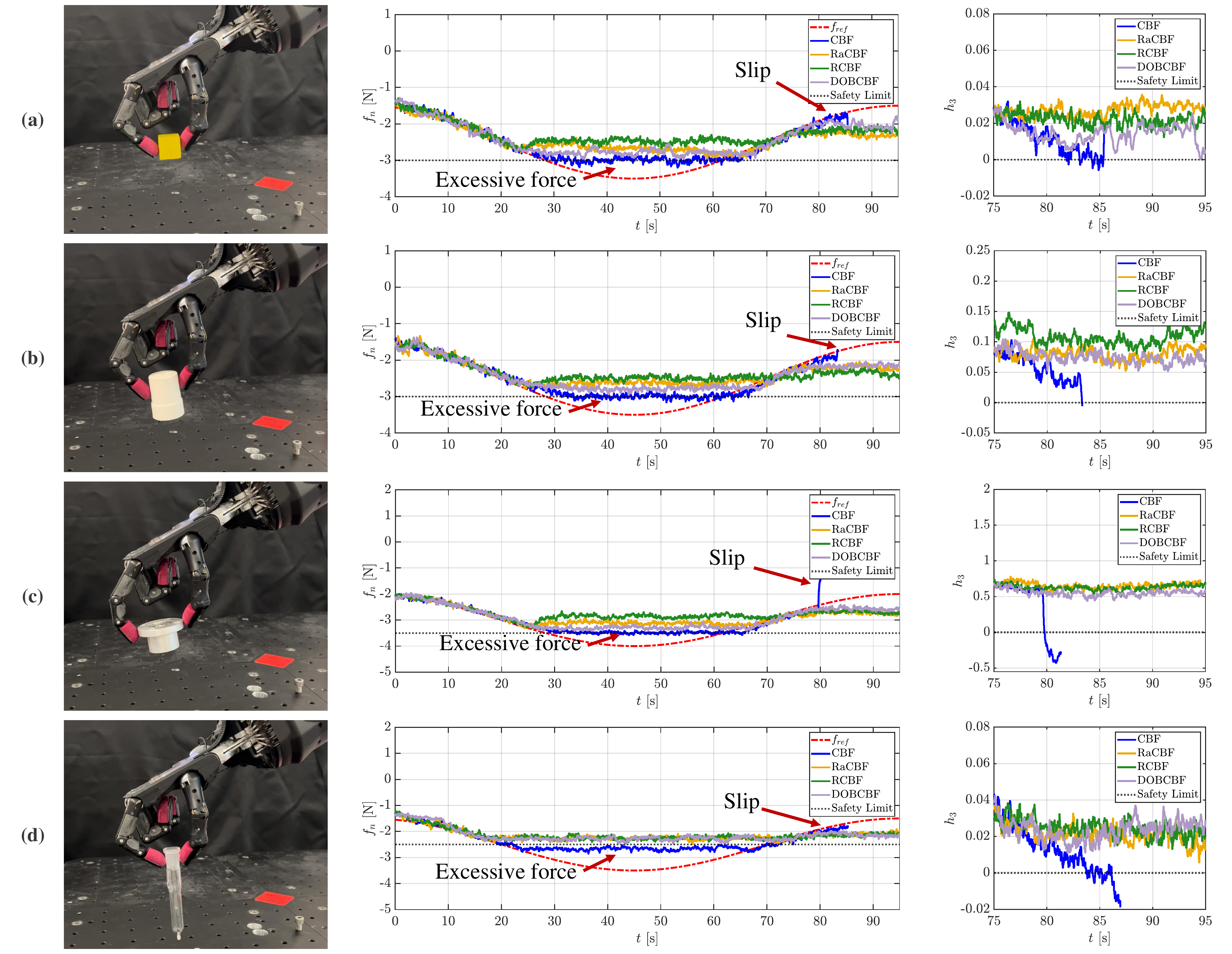}
    \caption{The experimental results of the safe grasping framework. We carry out various objects: (a) a cube, (b) a plastic cylinder, (c) an aluminium bearing housing, and (d) fragile lab glassware. The first column shows snapshots during grasping experiments, and the middle column presents the contact normal forces measured by tactile sensors on the front finger, and the last column shows the force closure constraint with $\hat{\mu}$.}
    \vspace{-0.35cm}
    \label{Exp:graphs}
\end{figure*}

With the use of a disturbance observer, the DOBCBF (purple) exhibits the least conservative behavior, since it estimates the model uncertainty as external disturbances and incorporates the estimated disturbances into its safety condition. Regarding the force closure constraint to avoid slippage, the standard CBF violates the constraint, resulting in slippage, as illustrated by the friction cone in Fig.~\ref{fig:sim_uncertainty} (c). However, the RCBF, RaCBF, and DOBCBF prevent the object from slipping and falling, staying inside the safety region indicated in pale green. Since the friction coefficient is accurately known in simulation, it is evident that the DOBCBF can reach the safety boundary while compensating for the external disturbances such as the model uncertainty.

Despite these positive results, we anticipate that determining the true friction coefficient in the real experiment is non-trivial unless we estimate it; thus, we assume that the friction coefficient error, $|\mu - \hat{\mu}|\leq e_{\textnormal{max}},$ $e_{\textnormal{max}} \geq 0$ is conservatively bounded as in \cite{Shaw2021} for the following real-world experiments.

\section{Hardware Experiments}\label{sec:experimental_validation}
In this section, we show the experimental results of the proposed method, demonstrating the safe grasping of various objects including a fragile object. To this end, we primarily use two fingers of the Shadow hand equipped with tactile sensors to grasp an object and control the front finger to regulate contact normal force. Lastly, we evaluate each safety filter in terms of safety violations and conservatism, and discuss the overall limitations of our framework. The demonstration video is available at the link\footnote{\url{https://youtu.be/Cuj47mkXRdg}}.

\begin{table*}[t] 
\centering
\centering
\begin{tabular}{cc|ccc|ccc|ccc|ccc}
\multicolumn{1}{c|}{\multirow{3}{*}{Setups}}                                       & Objects                     & \multicolumn{3}{c|}{Cube}                     & \multicolumn{3}{c|}{Plastic cylinder}        & \multicolumn{3}{c|}{Bearing housing}           & \multicolumn{3}{c}{Fragile lab glassware}            \\ \cline{2-14} 
\multicolumn{1}{c|}{}                                                                           & \multirow{2}{*}{Parameters} & \multicolumn{3}{c|}{$m=0.01$, $\hat{\mu} = 0.055$} & \multicolumn{3}{c|}{$m=0.03$, $\hat{\mu} = 0.17$} & \multicolumn{3}{c|}{$m=0.12$, $\hat{\mu} = 0.67$} & \multicolumn{3}{c}{$m=0.015$, $\hat{\mu} = 0.08$} \\
\multicolumn{1}{c|}{}                                                                           &                             & \multicolumn{3}{c|}{$\alpha = (80,65,65)$}      & \multicolumn{3}{c|}{$\alpha=(80,65,65)$}         & \multicolumn{3}{c|}{$\alpha=(80,100,100)$}         & \multicolumn{3}{c}{$\alpha=(90,70,70)$}         \\ \hline
\multicolumn{2}{c|}{Safety}                                                                                                   & $h_1$         & $h_2$         & $h_3$        & $h_1$          & $h_2$        & $h_3$        & $h_1$          & $h_2$        & $h_3$        & $h_1$         & $h_2$         & $h_3$       \\ \hline\hline
\multicolumn{1}{c|}{\multirow{4}{*}{\begin{tabular}[c]{@{}c@{}}Safety \\ Filters\end{tabular}}} & CBF \cite{Ames2019CBFtheoryandapplications}                         & \textcolor{red}{\textbf{-0.1983}}        & 0.1624        & \textcolor{red}{\textbf{-0.034}}       & \textcolor{red}{\textbf{-0.159}}        & 0.448       &\textcolor{red}{\textbf{-0.140}}       & \textcolor{red}{\textbf{-0.085}}         & 1.928     & \textcolor{red}{\textbf{-0.426}}        & \textcolor{red}{\textbf{-0.353}}      & 0.257       & \textcolor{red}{\textbf{-0.037}}      \\
\multicolumn{1}{c|}{}                                                                           & RaCBF \cite{lopez2020robust}                     & 0.049        & 0.214        & 0.018       & 0.195         & 0.641       & 0.0524       & 0.1871         & 2.898       & 0.544       & 0.01        & \textbf{0.3}        & \textbf{0.005}      \\
\multicolumn{1}{c|}{}                                                                           & RCBF \cite{JANKOVIC2018359}                      & 0.248        & 0.208        & 0.012       & 0.252         & 0.665       & 0.077       & 0.195         & 2.852       & 0.498       & 0.064      & 0.307        & 0.013      \\
\multicolumn{1}{c|}{}                                                                             & DOBCBF \cite{Wang2023}                     & \textbf{0.007}        & \textbf{0.196}        & \textbf{0.001}       & \textbf{0.046}         & \textbf{0.64}       & \textbf{0.0520}       & \textbf{0.078}        & \textbf{2.795}       & \textbf{0.4401}       & \textbf{0.003}        & 0.305        & 0.011      \\ \hline \hline
\end{tabular}

\caption{The performance of safety filters including CBF, RaCBF, RCBF, and DOBCBF for each experiment with different objects.}
\label{table_1}
\vspace{-0.7cm}
\end{table*} 

\subsection{Experimental Setup}
The proposed framework is developed and implemented in C++ via an ASUS FX507ZC4 laptop with i7-12700H processor, 16GB of RAM, and NVIDIA GeForce RTX 3050 GPU. The framework operates as a high-level controller at 125Hz while a low-level tracking controller runs at 1k Hz via an Intel NUC with an i7-10710U processor and 4GB of RAM. The implementation of quadratic programming in safety filters is carried out using QuadProg++ library \cite{Liu2021} which is based on \cite{Goldfarb1983}. The objects used in this experiment include a small wooden cube (10g), a plastic cylinder (30g), an aluminium bearing housing (120g), and a thin, fragile piece of lab glassware (15g) as shown in the first column of Fig.~\ref{Exp:graphs}. The friction coefficients $\hat{\mu}$ for each object are iteratively tuned to correspond to the minimum contact normal force required to grasp a given object without slippage. 

As shown in the middle column of Fig.~\ref{Exp:graphs}, we intentionally generate the desired force (the dashed red line) that violates safety constraints in order to evaluate the performance of each safety filter. For the safety constraints, we define a force regulation constraint for the contact normal force, $h_1$. The minimum normal reaction forces in this experiment are chosen as $-3$N, $-3$N, $-3.5$N, and $-2.5$N in the opposite direction to the applied force for the cube, the cylinder, the housing, and the glass object, respectively. For force closure constraints, we consider only a vertical cone plane with respect to the gravity direction, denoted by $h_2$ and $h_3$ across all experiments since the finger does not move in the horizontal direction at the contact point. Lastly, we use a linear function for an extended class $\mathcal{K}_\infty$ function, $\alpha(\cdot)$, (e.g., $\alpha h(\bm{x})$). The $\alpha$ values are the same for each safety filter, but vary for each object. For example, we conduct the experiment with a cube with $\alpha = (80,65,65)$ corresponding to each constraint, $h_1, h_2, h_3$, respectively. The rest of the experiment parameters are presented in Table~\ref{table_1}.

\subsection{Results and Discussion}
Figure~\ref{Exp:graphs}  presents the overall experimental results of the proposed framework. As shown in the middle column of Fig.~\ref{Exp:graphs} (a), the standard CBF (blue) failed to maintain force regulation and closure constraints due to the model uncertainty, resulting in excessive force applied to the object at first, followed by slippage due to insufficient grasping force at the end. We also observed that $h_3$ became negative as shown in the last column of Fig.~\ref{Exp:graphs} (a), indicating a safety violation. In contrast, the RCBF (green) showed significantly more robustness to uncertainty due to the maximum disturbance bound in the safety constraints. Likewise, we observe that the RaCBF is slightly robust to model uncertainty, but exhibits less conservative behavior than the RCBF since it adapts to the parametric uncertainty in the safety conditions. While RaCBF still has conservatism, the DOBCBF shows the least conservative performance since the estimated disturbances are directly attenuated in the safety conditions. Table~\ref{table_1} provides the minimum values of each safety condition for each safety filter, indicating that the CBF violates safety while other methods ensure safety. 

In the cases of the plastic cylinder, the housing, and the fragile glassware, we observe that the standard CBF violates safety, while the other methods maintain safety as shown in Fig.~\ref{Exp:graphs} (b)-(d) and Table~\ref{table_1}. As indicated in Table~\ref{table_1}, the conservative behaviors of each safety filter increasingly improve in the following order (with the exception of the glassware): RCBF, RaCBF, and DOBCBF. Since the feasible safe force region for the fragile object is much smaller than other objects, it is shown that the performance of each safety filter is similar in terms of conservatism. Compared to \cite{Shaw2021} using RCBF only, the proposed method leverages RaCBF and DOBCBF, thereby reducing conservatism by $83.5\%$ on average. 

Despite these positive results, numerous challenges were encountered in the hardware trials. First, there still exists uncertainty over $\mu$ even after iteratively tuning $\hat{\mu}$ to satisfy the force closure condition. To address this concern, we could employ friction coefficient estimation methods \cite{WU2024114249} from prior work for grasping to further reduce uncertainty in the force closure condition. Second, there is uncertainty in the contact frame on a grasped object. In the experiments, we assume that the contact frame is approximated as the finger's frame, which leads to model uncertainty in the system, deteriorating safety guarantees. However, these uncertainties can still be mitigated by using robust approaches such as RaCBFs, RCBFs or DOBCBFs, as demonstrated in the experiments. Another approach is to integrate in-hand pose estimation \cite{InHandPoseEstimation} to reduce the uncertainty for the contact frame, which can enhance the performance of the proposed framework.

\section{Conclusion}\label{sec:conclusions}
This paper presented a comprehensive framework for safe grasping with formal safety guarantees by using CBFs. We implemented three main components of the framework: safety filters, finger force controller, and the estimation of contact point and force/torque. We first designed contact force and force closure constraints and then enforced them into various safety filters to achieve safe grasping. We implemented and developed a finger force controller to track safe control input and a technique to estimate a contact point and force/torque from multiple tactile sensors at each finger. We simulated the framework in a scenario where two fingers grasp a cube, showing that the framework can achieve safe grasping with each safety filter. Finally, the proposed framework was experimentally validated with multiple objects including a fragile object on Shadow Robot setup, resulting in successful safe grasping without safety violations across all experiments. In the simulation and experiments, we also evaluated the performance of each safety filter in terms of safety violation and conservatism, and found that using DOBCBF is significantly advantageous for achieving safe grasping, since it provides the best performance to ensure safety guarantees with minimum conservatism.
\section*{Acknowledgements}
The work was supported by Fabrikant Vilhelm Pedersen og Hustrus Legat.
%%%%%%%%%%%%%%%%%%%%%%%%%%%%%%%%%%%%%%%%%%%%%%%%%%%%%%%%%%%%%%%%%%%%%%%%%%%%%%%%
%\section*{APPENDIX}
%\section*{ACKNOWLEDGMENT}
%%%%%%%%%%%%%%%%%%%%%%%%%%%%%%%%%%%%%%%%%%%%%%%%%%%%%%%%%%%%%%%%%%%%%%%%%%%%%%%%

\balance
\bibliographystyle{IEEEtran}
\bibliography{bibliography}
\end{document}